\pdfoutput=1

\documentclass[11pt]{article}

\usepackage[preprint]{acl}

\usepackage{times}
\usepackage{latexsym}

\usepackage[T1]{fontenc}

\usepackage[utf8]{inputenc}

\usepackage{microtype}

\usepackage{inconsolata}

\usepackage{graphicx}
\usepackage{algcompatible}
\usepackage{algorithm}
\usepackage{amsthm}
\usepackage{amsmath}
\usepackage{booktabs}
\usepackage{amssymb}
\usepackage{graphicx}
\usepackage{subcaption}
\usepackage{multirow}
\usepackage{tabularx}
\usepackage{diagbox}
\title{Advancing Multi-Party Dialogue Framework with Speaker-ware Contrastive Learning}

\author{
	\textbf{Zhongtian Hu\textsuperscript{1}},
	\textbf{Yiwen Cui\textsuperscript{1}},
	\textbf{Ronghan Li\textsuperscript{2}},
	\textbf{Meng Zhao\textsuperscript{3}},
	\textbf{Lifang Wang\textsuperscript{1}},
	\\
	\textsuperscript{1}School of Computer Science and Engineering, Northwestern Polytechnical University,
	\\\textsuperscript{2}School of Computer Science and Technology, Xidian University,
	\\\textsuperscript{3}School of Artificial Intelligence and Big Data, Henan University of Technology
	\\
	\texttt{ahxchzt@mail.nwpu.edu.cn},
	\texttt{wanglf@nwpu.edu.cn} 
}

\begin{document}
	
	\maketitle
\begin{abstract}
 Multi-party dialogues, common in collaborative scenarios like brainstorming sessions and negotiations, pose significant challenges due to their complexity and diverse speaker roles. Current methods often use graph neural networks to model dialogue context, capturing structural dynamics but heavily relying on annotated graph structures and overlooking individual speaking styles. To address these challenges, we propose \textbf{CMR}, a \textbf{C}ontrastive learning-based \textbf{M}ulti-party dialogue \textbf{R}esponse generation framework. CMR employs a \emph{two-stage self-supervised contrastive learning framework}. First, it captures global differences in speaking styles across individuals. Then, it focuses on intra-conversation comparisons to identify thematic transitions and contextually relevant facts. To the best of our knowledge, this is the first approach that applies contrastive learning in multi-party dialogue generation. Experimental results demonstrate that CMR not only significantly outperforms state-of-the-art models, but also generalizes well to large pre-trained language models, effectively enhancing their capability in handling multi-party conversations.
\end{abstract}
\section{Introduction}
\begin{figure}[!t]
	\centering
	\includegraphics[width=0.8\linewidth]{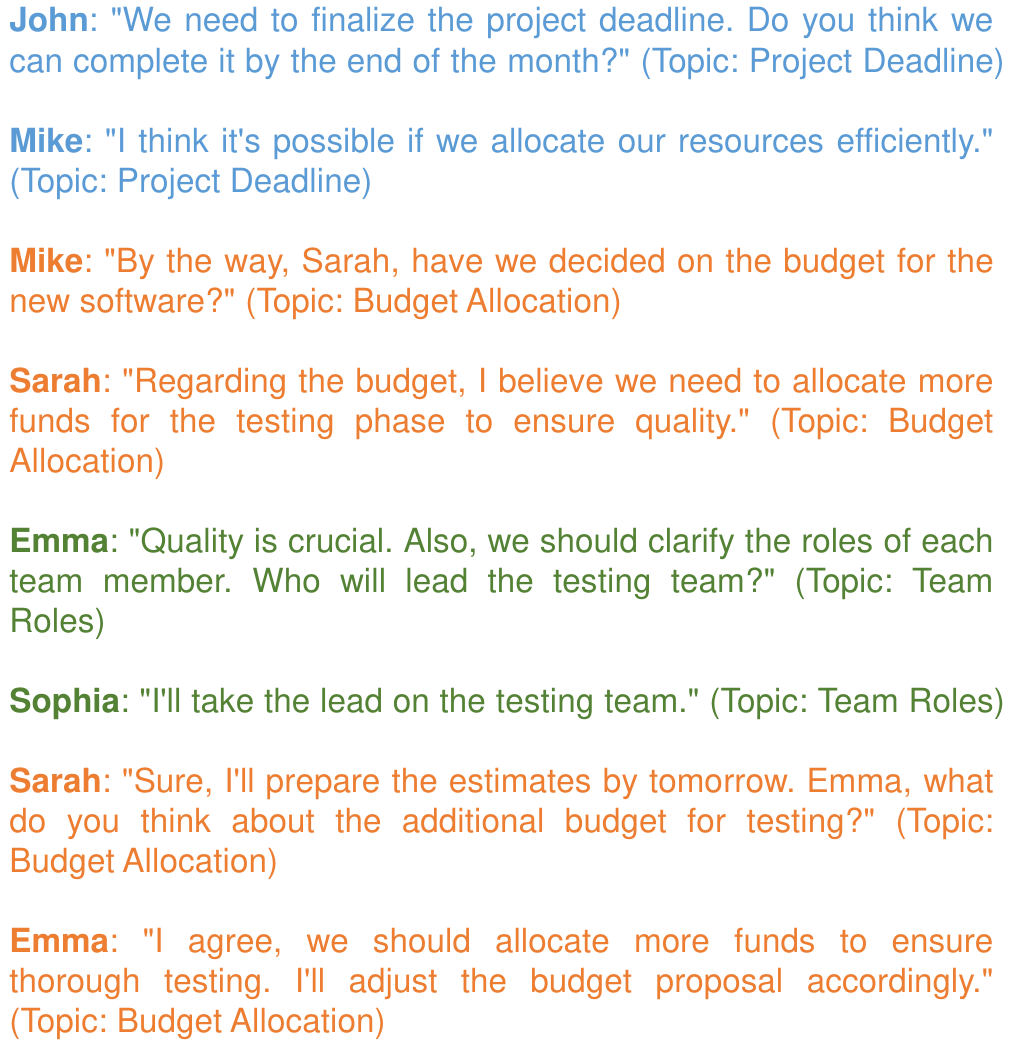}
	\caption{This figure illustrates a multi-party dialogue involving four participants: John, Mike, Sarah, and Emma. Each participant engages in discussions on different topics, demonstrating the dynamics and complexity of multi-party conversations.}
	\label{fig1}
\end{figure}

Dialogue response generation has made notable progress \cite{ide2021multi,LIU2021106} in applications \cite{hu2024dynamically, wu2022improving,komeili2022internet}, largely due to advancements in language models. However, most research has focused on dyadic dialogues \cite{cai2022pcvae,sun2021multimodal,chen2022dialogved}, involving only two participants. In contrast, multi-party dialogues, common in real-world scenarios such as group discussions, meetings, and role-play in game, present unique structural challenges. As shown in Figure \ref{fig1}, multi-party dialogues require careful management of information flow between multiple participants \cite{gu2022says,ganesh2023survey}.

To address these complexities, many approaches utilize graph neural networks (GNNs) to model dialogue context \cite{hu2019gsn, wang2020response, shi2019deep}. For example, \citet{gu2022hetermpc} propose the HeterMPC model, which uses utterance and interlocutor nodes with six meta relations to construct a heterogeneous graph that captures dialogue interactions. Similarly, \citet{gu2023madnet} introduce a fully connected graph to model utterance-interlocutor relationships, incorporating latent addressee inference. While GNNs effectively model the dependencies in multi-party dialogues, they suffer from \textbf{four major limitations}: \textbf{(1)} they require additional annotations such as speaker interaction labels; \textbf{(2)} they model each dialogue in isolation, failing to capture consistent speaker styles across conversations;  \textbf{(3)} they are insensitive to topic transitions, limiting their ability to track discourse dynamics. \textbf{(4)} their graph-based architectures are inherently difficult to scale or integrate into large pre-trained language models.

We propose \textbf{CMR} (Contrastive learning-based Multi-party Response Generation), a novel framework that leverages \textbf{self-supervised contrastive learning} to model the structure and dynamics of multi-party dialogues without relying on complex graph structures or extensive annotations. The model follows a two-stage training process. \textbf{Stage I} focuses on learning to differentiate speakers based on their utterances. By contrasting utterances from different speakers within the same conversation and those from random speakers across the dataset against the target speaker's utterances, the model learns to \textbf{distinguish speaking styles} among different participants. \textbf{Stage II} then optimizes the model for response generation with an integrated contrastive objective. Here, the model contrasts the dialogue context with generated responses (via beam search), the gold response, the target speaker's previous utterance, and other speakers' responses within the same conversation. This stage improves response quality by \textbf{enhancing the model's understanding of conversation themes and speaker-specific facts.}

We validate our approach on both the \textsc{Friends} dataset and the \textsc{Ubuntu} Dialogue Corpus. Specifically, \textbf{we evaluate CMR on two different backbone models}: a traditional encoder-decoder model (T5) and a decoder-only large language model (LLaMA 3.1). Evaluation results demonstrate that our model significantly outperforms the baseline models.

Our primary contributions are as follows:
\begin{itemize}
	\item We propose CMR, a novel multi-party dialogue response generation framework that leverages contrastive learning to better understand the structure and dynamics of multi-party dialogues. 
	
	\item We introduce a two-stage training process: the first stage focuses on distinguishing the speaking styles, and the second stage 
	improves the quality of responses and the model's understanding of dialogue themes and speaker-specific facts.
	
	\item  Evaluation results demonstrate that our model significantly outperforms baseline models and achieves SOTA.
\end{itemize}
\section{Related Work}
Research on multi-party dialogue response generation is relatively sparse, but there are some pioneering works in this field. Additionally, we review some methods that combine contrastive learning with generation tasks.
\subsection{Graph Neural Networks for Multi-Party Dialogue}
Several studies model multi-party dialogues using graph neural networks For instance, GSN (Graph-Structured Network) \cite{hu2019gsn} uses utterance-level graphs to capture dialogue flow and speaker dynamics. Similarly, HeterMPC \cite{gu2022hetermpc}, which employs heterogeneous graph networks to model interactions between interlocutors and utterances using multiple types of nodes and edges. Furthermore, MADNet \cite{gu2023madnet} utilizes fully connected graphs integrated with latent addressee inference via variational methods.
\subsection{Enhanced Speaker-Aware Approaches}
\cite{ma2023enhanced} introduce a model that applies a speaker mask  attention for comprehensive speaker-aware discourse clues.  \cite{liu2021filling} propose an Utterance-Aware and Speaker-Aware Multi-turn Dialogue Representation model (MDFN) that integrates utterance-level and speaker-level features using a mask attention mechanism. These methods highlight the importance of incorporating speaker-aware signals.
\subsection{Contrastive Learning for Text Generation}
Contrastive learning has been effectively applied to text generation tasks to enhance the quality of generated responses. For example, the work by \cite{an2022cont} on contrastive neural text generation demonstrates the benefits of using contrastive objectives to distinguish between similar and dissimilar contexts, thereby improving the coherence and relevance of generated text. Additionally, \cite{liu2021topic} introduced a topic-aware contrastive learning approach for abstractive dialogue summarization.
\section{Methodology}
\subsection{Task Formalization}
Formally, let $D = \{ ({u_1},{s_1}),...,({u_n},{s_m})\} $ represent a multi-party conversation, which contains $n$ utterances and $m$ speakers, and $u_i$ is the utterance made by speaker $s_j$. Our objective is to develop a model $f$ that can generate a response $r$ for a specific speaker $s_t$ based on the context of the dialogue $D$. This can be expressed as: $r=f(D,s_t)$.
\subsection{Model Overview}
\begin{figure*}[htbp]
	\centering
	\begin{subfigure}[t]{0.3\linewidth}
		\centering
		\includegraphics[width=\linewidth]{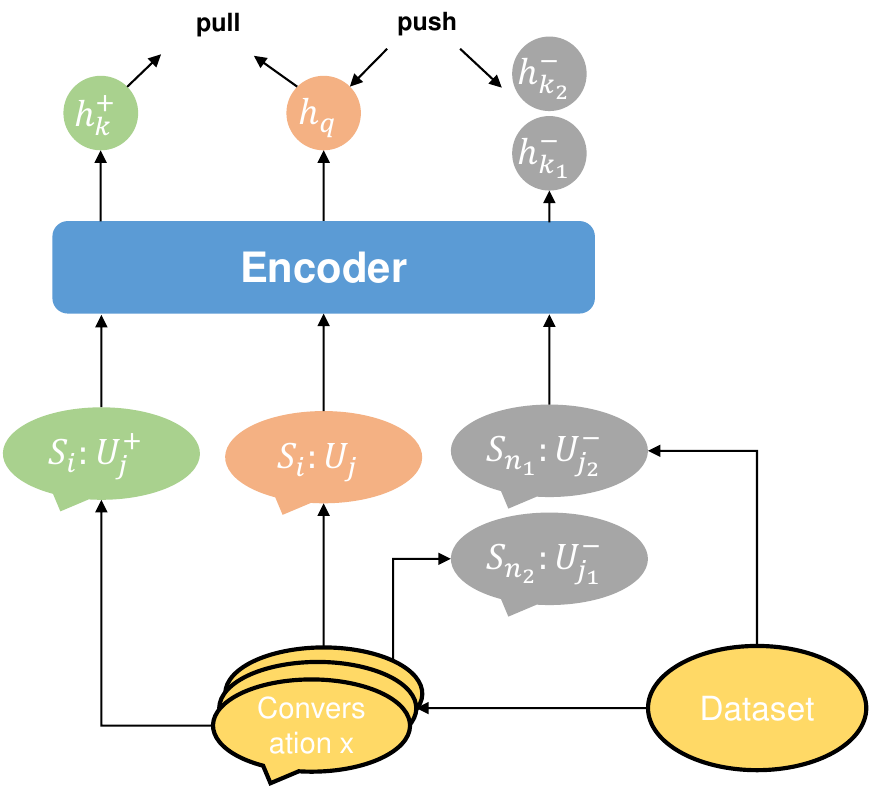} 
		\caption{The figure displays the training process of the Stage I.}
		\label{fig:sub1}
	\end{subfigure}
	\hspace{1cm}
	\begin{subfigure}[t]{0.3\linewidth}
		\centering
		\includegraphics[width=\linewidth]{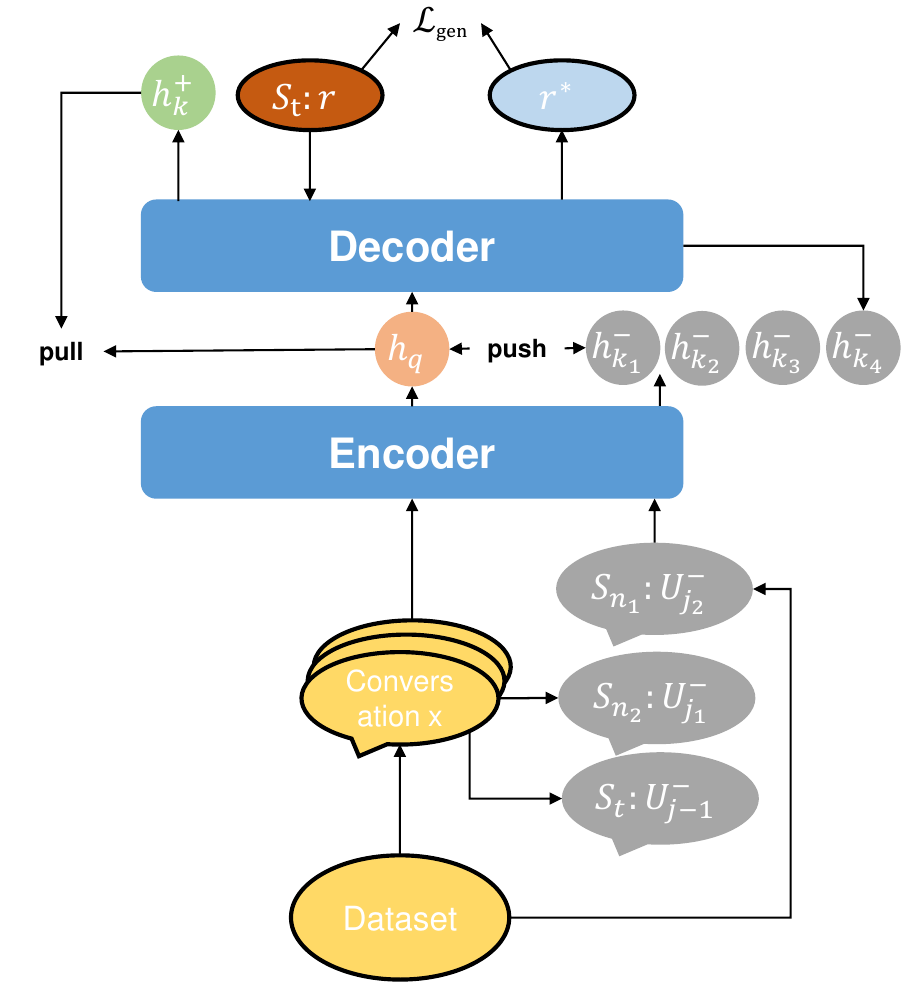} 
		\caption{The figure displays the training process of the Stage II.}
		\label{fig:sub2}
	\end{subfigure}
	\caption{The figure displays the training processes of the CMR framework. Here, an encoder-decoder architecture is presented as an example. Stage I (a) shows the encoder training with contrastive learning. Stage II (b) illustrates the joint training with response generation and contrastive learning.}
	\label{fig2}
\end{figure*}

The framework employs a two-stage training process. In the first stage, the model learns utterance-level representations using contrastive learning, enabling it to \textbf{distinguish between the speaking styles and contexts of different participants}. In the second stage, the model undergoes joint training with the response generation task and contrastive learning, enhancing its \textbf{understanding of dialogue themes and speaker-specific facts} to generate contextually relevant and accurate responses.

To provide a detailed understanding of the model's mechanics, we divide the discussion into two sections: Stage I (Sec \ref{sec3.3}) and Stage II(Sec \ref{sec3.4}). Each section covers the selection of positive and negative samples, as well as the training process specific to that stage.

\subsection{Stage I: Speaker Discrimination Learning}\label{sec3.3}
We use contrastive learning to differentiate speaker styles, contrasting utterances within and across conversation. 

\subsubsection{Stage I: Selection of Samples}
We select samples for contrastive learning as follows:
\begin{enumerate}
	\item \textbf{Query ($q$):} An utterance randomly selected from a conversation.
	\item \textbf{Positive Sample ($k^+$):} Another utterance spoken by the same speaker within the same conversation.
	\item \textbf{Negative Samples (${k^-}$):}
	\begin{itemize}
		\item Utterances spoken by other speakers within the same conversation.
		\item Utterances from random speakers across the dataset (in-batch samples).
	\end{itemize}
\end{enumerate}
It is worth noting that when selecting the query and positive and negative samples, we remove meaningless and dull replies, such as "yes yes yes," to ensure the quality and informativeness of the training data. The detailed methodology for data cleaning is provided in Appendix \ref{a}. 
\subsubsection{Stage I: Training Objective and Process}
We optimize the \textbf{InfoNCE} loss \cite{oord2018representation} to bring utterances from the same speaker closer while pushing apart those from different speakers, encouraging the model to better understand "who says what" and stylistic differences between different speakers:
\begin{equation}\label{eq1}
	\scriptsize
	\mathcal{L}_{\text {stage-1 }} =-log\frac{exp(\frac{h_q\cdot h_{k^+}}{\tau}) }{{\textstyle \sum_{i=1}^{N}}exp((\frac{h_q\cdot h_{k_i}}{\tau}) + {\textstyle \sum_{j=1}^{T}}exp(\frac{h_q\cdot h_{k^-_j}}{\tau})} 
\end{equation}
where the parameters $T$ and $N$ denote the number of negative samples from the same conversation and the batch size, respectively. $h_q$ represents the embedding of the query utterance, $h_{k^+}$ is the embedding of the positive sample, and $h_{k^-_j}$ are embeddings of negative samples drawn from the same conversation, $h_{k_i}$ represents the in-batch samples, and $\tau$ represents the temperature.

Specifically, for encoder-decoder architectures, these embeddings are obtained by encoding each utterance separately through the encoder and applying a pooling operation over the encoder's token-level hidden states. Thus, only encoder are trained for encoder-decoder model in Stage I. For decoder-only architectures, the embeddings are obtained by passing each utterance through the transformer-decoder individually and applying mean-pooling across the token-level hidden states produced by the decoder.
\subsection{Stage II: Enhancing Contextual Response}\label{sec3.4}
This stage involves joint training with response generation and contrastive learning, which aims to further refine the model's ability to generate contextually relevant and coherent responses.
\subsubsection{Stage II: Selection of Samples}
For a specific conversation, we select samples of the following types:
\begin{enumerate}
	\item \textbf{Query ($q$):} Dialogue context.
	\item \textbf{Positive Sample ($k^+$):} The gold response for the target speaker.
	\item \textbf{Negative Samples ($k^-$):} The negative samples \textbf{combine those from Stage I with additional types}:
	\begin{itemize}
		\item The target speaker’s previous utterance.
		\item Model-generated beam search samples.
	\end{itemize}
\end{enumerate}

\subsubsection{Stage II: Training Objective and Process}
For the response generation task, we use maximum likelihood estimation with cross-entropy loss. For the contrastive learning part, we use the InfoNCE loss, similar to Stage I. Thus, the training objective in Stage II combines the response generation loss and the contrastive learning loss. The combined loss function can be expressed as:
\begin{equation}
	\mathcal{L}_{\text {stage-2 }}=\mathcal{L}_{\text {gen }}+\lambda \mathcal{L}_{\text {cl}}
\end{equation}
where $\mathcal{L}_{\text {gen }}$ is the response generation loss, calculated using cross-entropy loss between the generated response and the gold response. $\mathcal{L}_{cl}$ is the contrastive learning loss, and 
$\lambda$ is a hyperparameter that balances the two loss components. 

The cross-entropy loss for response generation is defined as:
\begin{equation}
	\mathcal{L}_{\text {gen }}=-\sum_{t=1}^{T} \log P\left(r^{t} \mid r^{<t}, D, s_{t}\right)
\end{equation}
Here, $r^t$ represents the token at position $t$ in the gold response, $r^{<t}$ represents the sequence of tokens in the gold response before position $t$. $D$ is the dialogue context, and $s_t$ is the target speaker. The cross-entropy loss measures the negative log-likelihood of the gold response given the dialogue context and the target speaker.

The contrastive learning loss $\mathcal{L}_{cl}$ is defined as:
\begin{equation}
	\scriptsize
	\mathcal{L}_{cl} =-log\frac{exp(\frac{h_q\cdot h_{k^+}}{\tau}) }{exp(\frac{h_q\cdot h_{k^+}}{\tau}) + {\textstyle \sum_{j=1}^{P+T+N+B}}exp(\frac{h_q\cdot h_{k^-_j}}{\tau})} 
\end{equation}
The representations $h_q$ and $h_k$ for the query and samples are obtained in a manner similar to Stage I. The parameters $P, T, N, B$ denote the number of negative samples from four different sources.

The entire training process can be referenced in Appendix \ref{e}.
\section{Experiments}
\subsection{Dataset}
To evaluate the performance of CMR, we consider two widely-used benchmarks in multi-party dialogue research: the \textbf{Friends dataset}\cite{shmueli2019socialnlp} and the \textbf{Ubuntu IRC benchmark}\cite{lowe2015ubuntu}. The \textsc{Friends} dataset, which contains transcripts from the TV show "Friends," features interactions between six main characters with distinct and consistent personalities, providing a rich context for modeling speaking styles. In contrast, the \textsc{Ubuntu} dataset includes many speakers with sparse utterances, posing challenges for style modeling. Given these factors, \textbf{we determine that the \textsc{Friends} dataset is more suitable for evaluating our model's ability. However, to assess the generalization capability of our model, we apply the second-stage trained model to the \textsc{Ubuntu} dataset.}
\subsection{Comparison Models}
\begin{table*}[!t]
	\centering
		\resizebox{0.8\linewidth}{!}{
\begin{tabularx}{\textwidth}{l|>{\centering\arraybackslash}X>{\centering\arraybackslash}X>{\centering\arraybackslash}X>{\centering\arraybackslash}X>{\centering\arraybackslash}X}

			\toprule
			\diagbox{\textbf{Models}}{\textbf{Metrics}} 
			& \textbf{F1}   & \textbf{BLEU-1}  & \textbf{BLEU-2}  &\textbf{BLEU-3}   & \textbf{Rouge-L}    \\ \hline
						\multicolumn{6}{c}{\textbf{Friends}}  \\ \hline
			Bart\cite{lewis2020bart}              & 4.80   & 4.78  & 1.64  & 0.78   & 4.46        \\ 
			T5\cite{raffel2020exploring}         & 4.92   & 5.19  & 1.65  & 0.67   & 4.65         \\ 
			HeterMPC\cite{gu2022hetermpc}         &5.13  & 5.30     &  1.41 &  0.40   & 4.74    \\ 
			MADNet\cite{gu2023madnet}            & 6.17  & 6.00 &  1.54  & 0.63        & 5.95  \\  
			CONT\cite{an2022cont}         	  & 6.46  & 6.19 & 2.32     & 1.06          & 6.18   \\ 
			LLaMA3-8b-prompt         	  & 4.03  & 4.38 & 1.07     & 0.38  &3.75          \\ 
			LLaMA3-8b-ICT         	  & 4.91  & 4.92 & 1.57     & 0.79     &  4.64     \\ 
			LLaMA3-8b-lora         	  & 5.37  & 5.26 & 1.87     & 0.80     &  4.91     \\
			LLaMA3-70b-prompt         	  & 6.21  & 5.94 & 1.72     & 0.75   & 5.58        \\ 
			LLaMA3-70b-ICT         	  & 6.33  & 6.01 & 1.70     & 0.91     & 5.72      \\
			LLaMA3-70b-lora         	  & 5.33  & 5.12 & 2.11     & 0.98     &  4.90     \\ 
			\textbf{CMR-T5}         	&\textbf{8.14}   & \textbf{8.63} & \textbf{2.95}     & \textbf{1.39}          & \textbf{7.45}\\
			\textbf{CMR-LLaMA3 8b}         	&\textbf{8.71}   & \textbf{9.13} & \textbf{3.16}     & \textbf{1.44}          & \textbf{7.82}\\ \hline
			\multicolumn{6}{c}{\textbf{Ubuntu}}  \\ \hline
			GSN*\cite{hu2019gsn} & - &9.02 &2.78&-&-\\
			ASRG*\cite{9746498} &- &9.41 & 2.89 & -&-\\
			ChatMDG*\cite{LI2024102469} &- & 9.72 & 3.04 &- & -\\
						Bart\cite{lewis2020bart}              & 11.71   & 11.07  & 3.75  & 1.62   & 10.35        \\ 
			T5\cite{raffel2020exploring}         & 12.04   & 11.91  & 3.65  & 1.74   & 10.69         \\ 
			HeterMPC\cite{gu2022hetermpc}         &11.95  & 11.58     &  3.87 &  1.71   & 10.46    \\ 
			MADNet\cite{gu2023madnet}            & 12.13  & 11.64 &  3.81  & 1.78        & 11.13  \\  
			LLaMA3-8b-lora         	  & 12.75  & 12.39 & 3.98     & 1.80     &  12.07     \\
			LLaMA3-70b-lora         	  & 12.61  & 12.43 & 4.03     & 1.66     &  12.00     \\
			\textbf{CMR-T5(only Stage II)}         	&\textbf{12.41}   & \textbf{12.37} & \textbf{3.99}     & \textbf{1.80}          & \textbf{11.89}\\
			\textbf{CMR-LLaMA3 8b (only Stage II)}         	&\textbf{13.43}   & \textbf{13.01} & \textbf{4.28}     & \textbf{1.86}          & \textbf{12.89}\\
			\bottomrule
			
	\end{tabularx}}
	\caption{The table shows the performance of different models on various automatic evaluation metrics. For the specific settings of the LLaMA3 model, please refer to the appendix \ref{b}. * indicates that we directly report the results obtained from training on the original UBUNTU dataset as reported in the original paper.}
	\label{tab1}
\end{table*}
To evaluate the performance of CMR, we compare it against several \textbf{state-of-the-art} models for dialogue response generation, including both general-purpose and multi-party dialogue models. Evaluated models include graph-based methods (GSN\cite{hu2019gsn}, HeterMPC \cite{gu2022hetermpc}, MADNet \cite{gu2023madnet}, ChatMDG\cite{LI2024102469}), contrastive learning methods (CONT \cite{an2022cont}), and transformer-based models (BART \cite{lewis2020bart}, T5 \cite{raffel2020exploring}). Notably, \textbf{BART and T5 also serve as the backbone for all the aforementioned architectures.}  Lastly, we evaluate \textbf{LLaMA3.1}, a popular large language model. 
\subsection{Implementation Details}
Our CMR framework is combined with different backbone models, T5 and LLaMA 3.1. For LLaMA-based models, we apply LoRA during training to enable parameter-efficient fine-tuning. The learning rate is set to $4e^{-5}$ and we use the AdamW optimizer. In Stage I, the model is trained with contrastive learning for 10 epochs with a batch size of 4. For each batch, 7 negative samples are used, with 4 samples coming from the same conversation and 3 from other samples within the same batch. Stage II involves joint training with response generation and contrastive learning for 2 epochs, also with a batch size of 4 and $\lambda$ is setting to 2. In this stage, 8 negative samples are used, distributed as follows: 1 sample from the previous utterance by the target speaker, 2 samples from other speakers within the same conversation, 3 samples from the same batch, and 2 model-generated beam search samples. The temperature $\tau$ is set to 0.1.
\subsection{Automatic Evaluation Results}
As shown in Table \ref{tab1}, to evaluate the performance of our proposed CMR framework, we conduct experiments comparing it against several state-of-the-art models. The results are averaged over five runs.

On the \textsc{Friends} dataset, CMR demonstrates significant improvements across all evaluation metrics when combined with different backbone models, with several metrics nearly doubling. Specifically, compared to CONT, another contrastive learning-based dialogue response generation model, CMR-T5 substantially increases the F1 score by 26.02\%, and achieves notable gains on BLEU and ROUGE scores. Additionally, compared to graph-based models such as HeterMPC and MADNet, CMR-T5 also exhibits substantial performance improvements. Meanwhile, the decoder-only LLM variant, CMR-LLaMA, significantly outperforms the vanilla LLaMA model, further highlighting the effectiveness and generalizability of our approach.

For the \textsc{Ubuntu} dataset, due to its data format being incompatible with the full CMR model, we apply only \textbf{CMR (Stage II)} for evaluation. Despite this limitation, CMR still outperform all other models of the same scale.
\subsection{LLM-Based Pairwise Preference Evaluation}
We observe that for such complex multi-party dialogue scenarios, conventional automatic evaluation metrics may be insufficient to fully capture the quality differences between model responses. To address this, we employ the \textbf{GPT-4o model as an LLM-based judge}, instructing it to carefully consider aspects such as speaker style consistency, personalization, contextual coherence, and response informativeness.

\begin{figure}[!t]
	\centering
	\includegraphics[width=0.8\linewidth]{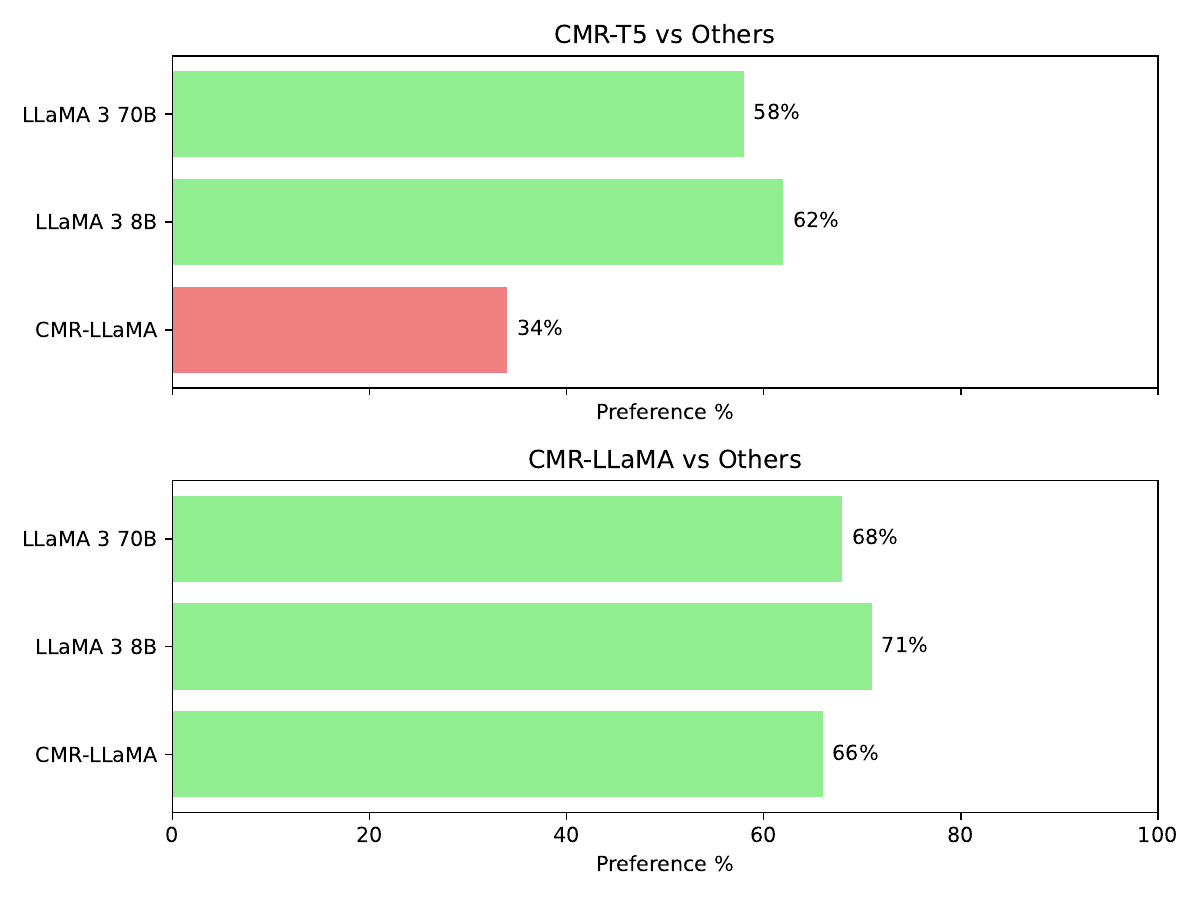}
	\caption{LLM-based pairwise preference comparison on the \textsc{Friends} dataset. }
	\label{fig:comparsion}
\end{figure}

Specifically, we randomly sample 500 dialogue instances from the \textsc{Friends} dataset and conduct pairwise preference comparisons among responses generated by CMR-T5, CMR-LLaMA, LLaMA 3.1 8B, and LLaMA 3.1 70B. As shown in Figure \ref{fig:comparsion}, although CMR-LLaMA achieves only slightly better scores than CMR-T5 on automatic metrics, the LLM-based preference evaluation shows that CMR-LLaMA is preferred in 66\% of the comparisons, indicating its superior ability to generate high-quality responses in multi-party dialogue settings.

\subsection{Human Evaluation Results}
In addition to automatic evaluation, we conduct human evaluations (Table \ref{tab2}) to assess the quality of responses generated by our CMR-T5 model compared to baseline models. We use four metrics for evaluation: \textbf{Relevance}, \textbf{Coherence}, \textbf{Engagement}, and \textbf{Speaker Appropriateness}. Here, Speaker Appropriateness checks if the response matches the target speaker’s style and persona. Annotators rate each response on a scale of 1 to 5, with dialogue contexts specifically selected to require responses from one of the six main characters in "Friends." Annotators, who are blind to the model that generated each response and were instructed to reviewing training set utterances to familiarize themselves with speaking style of the main characters.
\begin{table}[!t]
	\centering
	\resizebox{0.8\linewidth}{!}{
		\begin{tabular}{c|cccc}
			\toprule
			\diagbox{\textbf{Models}}{\textbf{Metrics}} 
			& \textbf{Rel.}   & \textbf{Coh.}  & \textbf{Eng.}  &\textbf{SA.}       \\ \hline
			Bart              & 1.42   & 3.01  & 1.23  & 1.25          \\ 
			T5          & 1.84   & 2.67  & 1.66  & 1.23            \\ 
			HeterMPC          &1.46   & 2.67     & 1.56  &1.52         \\ 
			MADNet            &2.12   &2.56  &2.13    & 1.87           \\  
			CONT-T5         	  & 2.61  & 3.21 & 3.04     & 2.27            \\ 
			\textbf{CMR}          	&\textbf{3.06}   & \textbf{3.40} & \textbf{3.09}     & \textbf{2.88}         \\ 
			\bottomrule
			
	\end{tabular}}
	\caption{The table shows the performance of different models on various human evaluation metrics. Our CMR model shows significant improvements over the baseline models.}
	\label{tab2}
\end{table}
\begin{table*}[!t]
	\centering
	\resizebox{0.6\linewidth}{!}{
		\begin{tabular}{c|ccccc}
			\toprule
			\diagbox{\textbf{Models}}{\textbf{Metrics}} 
			& \textbf{F1}   & \textbf{BLEU-1}  & \textbf{BLEU-2}  &\textbf{BLEU-3}   & \textbf{Rouge-L}    \\ \hline
			\textbf{CMR}          	&\textbf{8.14}   & \textbf{8.63} & \textbf{2.95}     & \textbf{1.39}          & \textbf{7.45}        \\ 
			-w/o SS1          & 7.37   & 7.64  & 2.55  & 1.26   & 6.77         \\ 
			-w/o SE1          &7.83  &7.99      &2.61   &1.31     & 7.02     \\ 
			-w/o Stage I      &7.17   &7.04  & 2.57    &1.21         &6.86   \\  
			-w/o PU          & 7.88   & 7.89  & 2.74  & 1.19   & 7.09         \\ 
			-w/o SS2          &6.93  &6.96      & 2.48  & 1.07     & 6.41    \\ 
			-w/o SE2            &7.05   &6.89  & 2.68   & 1.34        & 6.64  \\  
			-w/o BS            & 6.80  &6.52  &2.58    & 1.31        & 6.39  \\  
			-w/o Stage II &6.41   &6.32  & 2.24   & 1.04         & 5.91 \\
			\bottomrule
			
	\end{tabular}}
	\caption{The table shows the performance of the CMR model and its variants with specific components removed. }
	\label{tab3}
\end{table*}
As shown in Table 2, our model outperforms the baseline models in all metrics. Notably, general pre-trained models like BART and T5 often produce dull responses, with around 65\% of their outputs resembling generic phrases like "I am sorry" or "yes yes yes." This results in lower scores for relevance, engagement, and speaker appropriateness, despite relatively high coherence scores due to their logical consistency. Meanwhile, graph-based models like HeterMPC and MADNet struggle without additional annotations, with HeterMPC frequently generating meaningless phrases across all samples. Finally, while graph structures help improve relevance and coherence, particularly for HeterMPC and MADNet, contrastive learning in models like CMR proves more effective in capturing different speaking styles, demonstrating the unique advantages of each approach.
\section{Analysis and Discussion}\label{sec5}
All analyses in the following sections are conducted based on the CMR-T5 model.
\subsection{Ablation Study}
To understand the contribution of each component of our CMR model, we conduct a series of ablation experiments. Each ablation removes a specific component or modifies the model's training process, allowing us to observe the impact on performance. As mentioned earlier, we conducted ablation experiments on \textsc{Friends} to isolate component impacts, as \textsc{Ubuntu}'s format limits full CMR application. The results of these ablation studies are summarized in Table \ref{tab3}.
\subsubsection{Impact of SS}
The removal of SS1 (same conversation's other speakers' utterances as negative samples in Stage 1) and SS2 (same conversation's other speakers' utterances as negative samples in Stage 2) both resulted in performance drops across all metrics. Specifically, the F1 score decreased by 0.77 points when SS1 was removed and by 1.21 points when SS2 was removed. These results indicate the importance of incorporating context-specific negative samples to enhance the model's ability to distinguish between relevant and irrelevant utterances.
\subsubsection{Impact of Stage I}
Removing the entire Stage I results in a significant performance drop across all metrics. The F1 score decreased by 0.97 points, BLEU-1 by 1.59 points, BLEU-2 by 0.38 points, BLEU-3 by 0.18 points, and ROUGE-L by 0.59 points. This demonstrates that the contrastive learning stage is crucial for improving the model's understanding of dialogue context and speaker styles.
\subsubsection{Impact of PU and BS}
In Stage 2, removing PU (previous utterance by the target speaker as a negative sample) and BS (beam search generated utterances as negative samples) led to noticeable declines in performance. These results suggest that PU helps the model achieve a fine-grained understanding of the dialogue context, while BS significantly enhances the overall quality of the generated responses.
\subsubsection{Impact of Stage II}
Removing the entire Stage II resulted in the most significant performance drop across all metrics. The F1 score decreased by 1.73 points, BLEU-1 by 2.31 points, BLEU-2 by 0.71 points, BLEU-3 by 0.35 points, and ROUGE-L by 1.54 points. This underscores the critical role of the joint training phase in fine-tuning the model's response generation capabilities. Without Stage II, the model struggles to maintain the coherence and relevance of responses, demonstrating the necessity of this stage for achieving high-quality dialogue generation.
\subsection{Impact of the Number of Negative Samples}
The number of negative samples is crucial for contrastive learning. Therefore, we explored the impact of varying the number of negative samples on the performance of CMR. As shown in Appendix \ref{c}.
\subsection{Visualization of the Impact of Speaker aware in Stage I}
To further understand the impact of Speaker aware in Stage I on the performance of the CMR-T5 model, we visualize the representations learned by the model with and without Stage I, as shown in Appendix \ref{d}.
%
\subsection{Generalization and Application Scenarios.}
CMR is particularly well-suited for scenarios where a fixed set of speakers appears across dialogues, such as NPC conversations in games or episodic scripted content. In these cases, modeling consistent speaker styles across samples can bring notable benefits. 

Nonetheless, CMR is not restricted to such settings. On the \textsc{Ubuntu} dataset where speakers rarely repeat, CMR still outperforms all baseline models of the same scale, despite being slightly behind LLM. We can say that the second stage of our framework alone is effective in modeling multi-party dialogues, even in the absence of speaker-level recurrence. Moreover, unlike prior methods, CMR requires no additional annotations, further demonstrating its adaptability to diverse dialogue scenarios.

\textbf{To further evaluate the generalization ability of CMR, we conduct additional experiments in settings where the model encounters speakers with limited presence in the training data.} As detailed in the appendix \ref{appendixf}, CMR still achieves strong performance under this condition. Intuitively, the phenomenon may be attributed to the model's ability to generalize from frequently seen speakers. While the model may be unable to fully capture the speaking style of rarely seen speakers, it can effectively leverage its understanding of other speakers to distinguish between interlocutors. Thus, CMR remains capable of identifying speaker-specific factual content and maintaining consistency in speaker-related information.

\subsection{Isolating Speaker Style: Comparison with Per-Speaker Models}
One of the core motivations of CMR is to model individualized speaker style within multi-party dialogues. A natural question arises: would training a separate model for each speaker yield better style modeling, as each model can specialize in the language patterns of a single character?

To address this question, we conduct an experiment on the Friends dataset by training six separate models, one for each of the six main characters. During inference, we invoke the model corresponding to the ground-truth speaker. Please note that although the Friends dataset includes additional speakers beyond the six main ones, this experiment is restricted to training and testing only on the six main characters. As such, the results of CMR in this setting are slightly different from those reported earlier. The details are shown in Appendix \ref{g}, the experimental results clearly demonstrate that training separate models for each speakers significantly poorer performance compared to CMR.
\subsection{Robustness to Varying Speaker Counts}

To evaluate the robustness of our model in handling dialogues with different levels of speaker complexity, we conducted additional experiments on subsets of dialogues containing 3, 4, 5, and 6 speakers.

\paragraph{Using the Entire Dataset.}
We first tested our model, trained on the entire Friends dataset, on conversations with varying speaker counts. We observe a gradual performance decrease as the number of speakers increases. \textbf{However, the decline may be attributed not only to the inherent complexity of multi-party dialogue but also to the training data distribution}, which predominantly features conversations with 3 or 4 speakers.

\paragraph{Using Balanced Subsets.}
To isolate the effect of speaker count, we created balanced subsets of 400 samples for each speaker setting. We retrained the model on these balanced subsets. Although the overall scores are lower compared to the model trained on the full dataset, \textbf{the performance decline across speaker counts is notably smaller}, suggesting that our model retains a reasonable degree of robustness when faced with increasing dialogue complexity.

All details and results are shown in Appendix \ref{h}.

\subsection{Training Cost and Deployment}

Due to CMR's two-stage training strategy and the need to generate responses for contrastive learning during the second stage, CMR requires more training time than its backbone model.

However, once training is complete, CMR demonstrates a distinct advantage over graph-based models during deployment. The contrastive learning component of CMR is entirely absent during inference, meaning that its resource requirements and inference speed are identical to those of a standard T5 model. In contrast, graph-based models require the construction of graphs during both training and inference, which incurs additional computational overhead. 

\section{Conclusion}
We propose CMR, a novel contrastive learning-based multi-party dialogue response generation framework. Our approach leverages a two-stage training process to enhance the model's understanding of dialogue context and speaker styles. In Stage I, we employ contrastive learning to differentiate between relevant and irrelevant utterances, significantly improving the model's ability to comprehend complex multi-party dialogues. Stage II combines response generation with contrastive learning, further refining the model's capability to produce contextually relevant and coherent responses. In conclusion. Our CMR framework advances multi-party dialogue response generation by integrating contrastive learning to effectively handle complex dialogue dynamics and speaker variability, providing a robust foundation for future research in this domain.
\section{Limitations}
While the CMR model demonstrates strong performance in multi-party dialogue response generation, several limitations must be acknowledged. 
\begin{itemize}
\item While the model does not require additional resources compared to the backbone model during deployment, its training process is time-consuming due to the multi-stage contrastive learning framework.
\item While the model effectively captures speaker-specific characteristics in datasets with distinct characters, such as \textsc{Friends}, and also demonstrates promising results on \textsc{Ubuntu} by leveraging only Stage II training, it may still be inherently better suited for scenarios where the set of speakers is relatively fixed and speaker-specific features are explicitly available.

\end{itemize}
	%

\bibliography{custom}

\appendix

\section{The algorithm of entire training process}\label{e}
See Algorithm \ref{al1}.
\begin{algorithm}[h]
	\caption{Training CMR Model}
	\begin{algorithmic}[1]	
		\STATE \textbf{Stage I: Encoder Training}
		\FOR{epoch in num\_epochs\_stage\_1}
		\FOR{batch in dataloader}
		\STATE Initialize $loss_{stage\_1} \leftarrow 0$
		\FOR{dialogue, speaker in batch}
		\STATE Select query $q$ from dialogue
		\STATE Select positive sample from same speaker
		\STATE Select negative samples from other speakers and batch
		\STATE Compute InfoNCE loss
		\STATE $loss_{stage\_1} \leftarrow   loss_{stage\_1} + loss_{InfoNCE}$
		\ENDFOR
		\STATE Backpropagation and optimization
		\ENDFOR
		\ENDFOR
		\STATE \textbf{Stage II: Joint Training}
		\FOR{epoch in num\_epochs\_stage\_2}
		\FOR{batch in dataloader}
		\STATE Initialize $loss_{gen} \leftarrow 0$, $loss_{cl} \leftarrow 0$
		\FOR{dialogue D, target speaker $s_t$ in batch}
		\STATE Generate response $\hat{r}$ and negative samples from beam search
		\STATE Get gold response $r$
		\STATE Compute generation loss: $loss_{gen} \leftarrow loss_{gen} + \mathcal{L}_{gen}$
		\STATE Select query $q \leftarrow D$
		\STATE Select positive sample $r$
		\STATE Select negative samples from previous utterance, other speakers, batch, beam search
		\STATE Compute InfoNCE loss: $loss_{cl} \leftarrow loss_{cl} + \mathcal{L}_{cl}$
		\ENDFOR
		\STATE Combine losses: $loss_{stage\_2} \leftarrow loss_{gen} + \lambda \cdot loss_{cl}$
		\STATE Backpropagation and optimization
		\ENDFOR
		\ENDFOR		
	\end{algorithmic}  		\label{al1}
\end{algorithm}
\section{Clean Dataset}\label{a}

The \textsc{Friends} and \textsc{Ubuntu} dataset we obtained included many low-quality samples. Some samples contained numerous meaningless and repetitive phrases, while others were too short and involved only two speakers. Additionally, some samples were fragmented and lacked context, making them unsuitable for training a robust dialogue model.
\begin{table}[h]
	\centering
	\begin{tabular}{|l|l|l|l|}
		
		\multicolumn{4}{c}{\textbf{Bad Words}}  \\ \hline
		yes                & okay      & hey       & oh        \\ \hline
		no                 & hello     & yeah      & well      \\ \hline
		aha                & ok        & i         & am        \\ \hline
		fine               & too       & bye       & is        \\ \hline
		good               & hm        & alright   & really    \\ \hline
		you                & are       & ohh       & uh        \\ \hline
		huh                & ha        & what      & go        \\ \hline
		hi                 & thanks    & right     & many      \\ \hline
		and                & ohhh      & could     & can       \\ \hline
		that               &           &           &           \\ \hline
	\end{tabular}
	\caption{List of Bad Words}
	\label{table:bad_words}
\end{table}

\begin{table*}[!t]
	\centering
	\resizebox{0.8\linewidth}{!}{
		\begin{tabular}{c|cccccc}
			\toprule
			\diagbox{\textbf{$N_1,T_1,N_2,T_2,P,B$}}{\textbf{Metrics}} 
			& \textbf{F1}   & \textbf{BLEU-1}  & \textbf{BLEU-2}  &\textbf{BLEU-3}   & \textbf{Rouge-L}    \\ \hline
			$1,1,1,1,1,1$         	&7.84   & 8.00 & 2.63     & 1.40          & 7.09        \\
			$2,2,2,1,1,1$         	&7.95   & 8.47 & 2.69     & 1.37          & 7.20        \\ 
			$3,3,3,1,1,1$          & 7.96   & 8.44  & 2.76  & 1.40   & 7.19         \\ 
			$\mathbf{4,4,4,2,1,2}$          	&\textbf{8.14}   & \textbf{8.63} & \textbf{2.95}     & \textbf{1.39}          & \textbf{7.45}     \\ 
			$5,5,5,3,1,3$      &8.17   &8.64  & 2.87    &1.41         &7.51   \\  
			$6,6,6,3,2,3$          & 8.20   & 8.76  & 2.89  & 1.37   & 7.54         \\ 
			$8,8,8,4,3,4$          &8.02  &8.58      & 2.77  & 1.34     & 7.20    \\ 
			\bottomrule
			
	\end{tabular}}
	\caption{Performance of the CMR model with different numbers of negative samples ($N_1$, $T_1$, $N_2$, $T_2$, $P$, $B$). Increasing the number of negative samples initially improves performance, but beyond a certain point, the results start to fluctuate, indicating that excessively high numbers of negative samples can introduce irrelevant or low-quality negatives, which can degrade the model's effectiveness.}
	\label{tab4}
\end{table*}
\begin{table}[!htbp]
	\begin{tabularx}{\linewidth}{l|l|l|l}
		\hline
		\multirow{4}{*}{\textbf{Prompt}}               & \multicolumn{3}{l}{$X_t$. You participated in }\\
		& \multicolumn{3}{l}{this conversation as $S_t$. }        \\ 
		&\multicolumn{3}{l}{Please provide your } 	\\
		&\multicolumn{3}{l}{response. } \\ \hline
		\textbf{In-context Learning}               & \multicolumn{3}{l}{$X_i$. $S_i$ said: $Y_i$ ... $X_t$ }        \\ \hline
	\end{tabularx}
	\caption{Configurations for LLaMA-3 using Prompt and In-context Learning. $t$ represents the target sample, and $i$ represents a random sample from the training set.}
	\label{table:llama}
\end{table}
To address these issues, we first set a list of ``bad words," as shown in Table \ref{table:bad_words}. If a sample's response contained these bad words in more than 30\% of its content, the sample was removed. Additionally, we removed samples where the dialogue involved only two speakers. Finally, we manually removed samples that lacked sufficient context. After data pre-processing, there are 5000 conversations for training and 653 conversations for testing on \textsc{Friends} dataset. For the \textsc{Ubuntu} dataset, since each utterance is relatively short, we also require that each dialogue contains at least 6 utterances and the entire conversation needed to exceed 512 tokens. There are about 67000 conversations for \textsc{Ubuntu} dataset. 

\section{Configuration of LLaMA}\label{b}
We use LLaMA 3.1 as our comparison model. For the configurations of LLaMA-3 using Prompt and in-context learning, refer to Table \ref{table:llama}. In our experiments, the prompt shown in the table yielded the best results and is used in the reported experiments. The in-context learning configuration involved testing with 1 to 6 samples, and we found that using 3 samples provided the best performance.

\section{Impact of the Number of Negative Samples}\label{c}
As shown in Table \ref{tab4}, increasing the number of negative samples initially improves the experimental results. However, beyond a certain point, the performance begins to fluctuate and does not continue to improve steadily. This fluctuation may be due to the inclusion of more irrelevant or less similar negative samples as the number of negatives increases. For instance, as the value of k increases in beam search-generated top-k negative samples, more less optimal negative samples are included, which can negatively impact the effectiveness of contrastive learning.
\section{Visualization of the Impact of Speaker aware in Stage I}\label{d}
\begin{figure}[htbp]
	\centering
	\begin{subfigure}[t]{\linewidth}
		\centering
		\includegraphics[width=\linewidth]{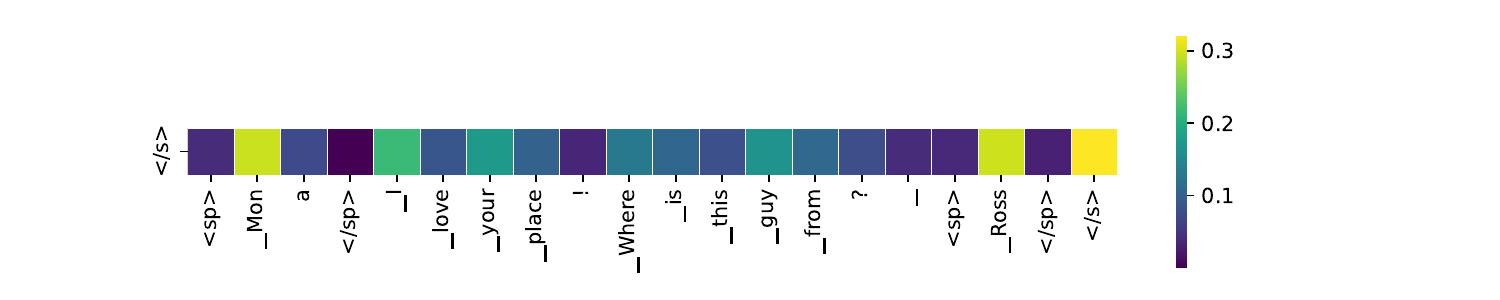} 
		\caption{}
		\label{fig3:sub1}
	\end{subfigure}
	\hspace{1cm}
	\begin{subfigure}[t]{\linewidth}
		\centering
		\includegraphics[width=\linewidth]{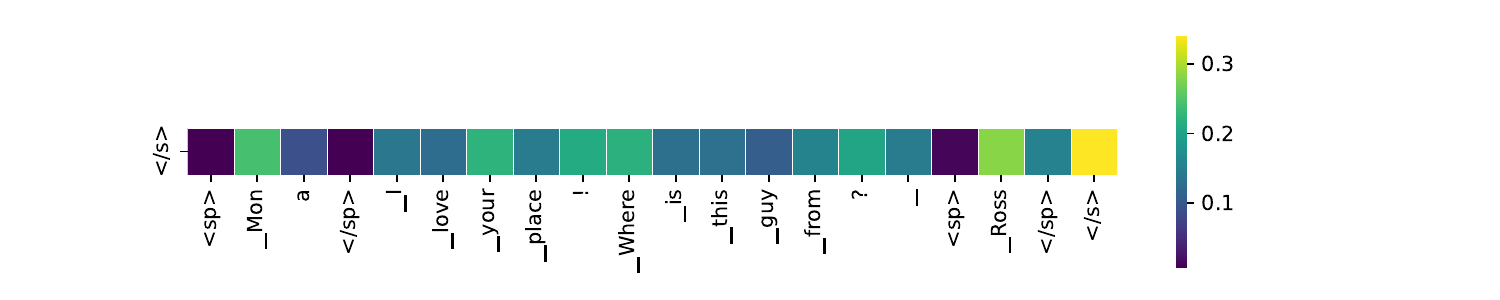} 
		\caption{}
		\label{fig3:sub2}
	\end{subfigure}
	\caption{Attention weights of the $<$/s$>$ token in the encoder for (a) the CMR model with Stage I contrastive learning, and (b) the CMR model without Stage I contrastive learning. The attention weights are more focused on contextually relevant tokens such as names and pronouns in (a), indicating improved context understanding after Stage I.}
	\label{fig3}
\end{figure}
As shown in Figure \ref{fig3}, we visualize the attention weights of the $<$/s$>$ token to other tokens in the encoder. The reason for choosing the $<$/s$>$ token, as mentioned earlier, is based on our previous experiments. We found that during generation, the cross-attention weights of the decoder in the T5 model are primarily concentrated on the $<$/s$>$ token in the encoder. This indicates that the $<$/s$>$ token acts as an information aggregator in encoder, and its attention weights to other context tokens can effectively reflect the model's primary focus. As shown in Figure 2, the attention weights of the $<$/s$>$ token in the CMR model with Stage I show a more concentrated pattern. The attention is focused on specific tokens that are contextually relevant, such as names (``Mona", ``Ross") and pronouns (``I," ``you"), indicating a better understanding of the dialogue context.
\section{Generalization to Rare and Unseen Speakers}\label{appendixf}
To investigate the generalization capabilities of CMR with respect to rarely observed or completely unseen speakers during training, we extract dialogues from the \textsc{Friends} dataset involving speakers beyond the six main characters. Specifically, we isolate response samples from the test set where the speakers are infrequently encountered or entirely absent in the training phase. These samples are evaluated using the fully trained CMR model without any further fine-tuning and the results in Table \ref{table:rare_unseen_speakers} illustrate the performance of CMR on responses by rare or unseen speakers.

\begin{table}[h]
	\centering
	\resizebox{\linewidth}{!}{
	\begin{tabular}{lccccc}
		\hline
		\textbf{Evaluation} & \textbf{F1} & \textbf{BLEU-1} & \textbf{BLEU-2} & \textbf{BLEU-3} & \textbf{ROUGE-L} \\
		\hline
		\textbf{Rare Speakers} & 8.11 & 8.65 & 2.85 & 1.38 & 7.29 \\
		\textbf{Standard}         	&8.14   & 8.63 & 2.95     & 1.39          & 7.45\\ 
		\bottomrule
	\end{tabular}}
		\caption{CMR Performance on responses from rare and unseen speakers.}
	\label{table:rare_unseen_speakers}
\end{table}

The experimental outcomes demonstrate that the CMR model maintains strong performance in generating responses even for speakers rarely or never encountered during training.

A plausible explanation for this generalization ability is that although the CMR model may not fully capture the speaking style of rarely seen or unseen speakers due to limited data exposure, it can effectively leverage the comprehensive understanding gained from frequent speakers. With the assistance of the first stage, while the model may be unable to fully capture the speaking style of rarely seen speakers, it can effectively leverage its understanding of other speakers to distinguish between interlocutors. The second stage, is less affected by unseen speakers because it primarily focuses on dialogue themes, context continuity, and factual consistency. Consequently, CMR remains capable of recognizing thematic shifts and accurately maintaining speaker-related factual content, thereby ensuring coherent and contextually relevant responses despite encountering rarely or never-before-seen speakers.
\section{Training separate models for each speakers}\label{g}
We extract samples for the six main characters from the Friends dataset and train six baseline models separately. During the testing phase, a different baseline model will be invoked according to the speaker to evaluate its performance.
\begin{table}[h]
	\centering
	\resizebox{\linewidth}{!}{
	\begin{tabular}{lccccc}
		\hline
		\textbf{Method} & \textbf{F1 Score} & \textbf{BLEU-1} & \textbf{BLEU-2} & \textbf{BLEU-3} & \textbf{ROUGE-L} \\
		\hline
		Separate models & 7.33 & 7.90 & 2.61 & 1.25 & 7.04 \\
		CMR            & 8.14 & 8.63 & 2.95 & 1.39 & 7.45 \\
		\hline
	\end{tabular}}
	\caption{Comparison between separate models and our proposed method.}
	\label{table:separate_vs_ours}
\end{table}
The experimental results clearly demonstrate significantly poorer performance compared to CMR. We attribute this outcome to two primary factors:
\begin{itemize}
\item The approach of training separate models for each character only enables the model to capture superficial personalized information about individual characters. More importantly, it fails to help the model understand the nuanced differences between different characters.

\item The original Friends dataset has a limited number of samples to begin with. After splitting the training process, each model only has access to approximately one-sixth of the original data.
\end{itemize}
\section{Robustness to Varying Speaker Counts}\label{h}
In this section, we provide additional details and full results for our robustness analysis on varying speaker counts.

\paragraph{Experimental Setup.}
We conducted two sets of experiments to assess the model’s performance when handling conversations with different numbers of participants (3, 4, 5, and 6 speakers):

\begin{itemize}
	\item \textbf{Experiment 1: Using the Entire Dataset.} We evaluated the model trained on the full dataset without controlling for speaker count distribution. This reflects a realistic but potentially biased setting, as most training samples involve only 3 or 4 speakers.
	
	\item \textbf{Experiment 2: Using Balanced Subsets.} To control for data imbalance, we created balanced subsets with 400 samples each for conversations involving 3, 4, 5, and 6 speakers. We retrained the model on each subset and evaluated its performance.
\end{itemize}

\paragraph{Results on the Entire Dataset.}
\begin{table}[h]
	\centering
	\resizebox{\linewidth}{!}{
	\begin{tabular}{c|ccccc}
		\hline
		\textbf{\# Speakers} & \textbf{F1} & \textbf{BLEU-1} & \textbf{BLEU-2} & \textbf{BLEU-3} & \textbf{ROUGE-L} \\ \hline
		3 & 8.50 & 9.00 & 3.10 & 1.50 & 7.80 \\
		4 & 8.21 & 9.13 & 2.99 & 1.47 & 7.49 \\
		5 & 7.91 & 8.47 & 2.88 & 1.35 & 7.36 \\
		6 & 7.65 & 8.22 & 2.73 & 1.30 & 7.10 \\ \hline
	\end{tabular}}
	\caption{Performance on conversations with 3--6 speakers using the entire dataset.}
	\label{tab:speaker_analysis_full}
\end{table}

\paragraph{Results on Balanced Subsets.}
\begin{table}[h]
	\centering
	\resizebox{\linewidth}{!}{
	\begin{tabular}{c|ccccc}
		\hline
		\textbf{\# Speakers} & \textbf{F1} & \textbf{BLEU-1} & \textbf{BLEU-2} & \textbf{BLEU-3} & \textbf{ROUGE-L} \\ \hline
		3 & 7.84 & 8.46 & 2.93 & 1.48 & 7.28 \\
		4 & 7.69 & 8.23 & 2.82 & 1.41 & 7.12 \\
		5 & 7.69 & 8.13 & 2.77 & 1.39 & 7.09 \\
		6 & 7.63 & 8.22 & 2.73 & 1.37 & 7.11 \\ \hline
	\end{tabular}}
	\caption{Performance on balanced subsets with 3--6 speakers (400 samples each).}
	\label{tab:speaker_analysis_balanced}
\end{table}

\paragraph{Analysis.}
The results across both experimental settings reveal a consistent performance drop as the number of speakers increases. However, when using balanced subsets, the performance degradation is much less pronounced, suggesting that our model can generalize reasonably well across varying

\end{document}